\documentclass[a4paper,fleqn]{cas-dc}

\usepackage[authoryear]{natbib}
\usepackage[utf8]{inputenc}
\usepackage{times}
\usepackage{epsfig}
\usepackage{graphicx}
\usepackage{amsmath}
\usepackage{amssymb}
\usepackage{color}
\usepackage{multirow}
\usepackage{algorithmic}
\usepackage{algorithm}
\usepackage{bbding} 
\usepackage{array}
\usepackage{textcomp}
\usepackage{stfloats}
\usepackage{url}
\usepackage{verbatim}
\usepackage{graphicx}
\usepackage{tablefootnote}
\usepackage{ulem} 
\usepackage{mathrsfs}
\usepackage{ulem}
\usepackage[switch]{lineno}

\makeatletter
\renewcommand{\fnum@figure}{Fig. \thefigure.\@gobble}
\makeatother

\def\tsc#1{\csdef{#1}{\textsc{\lowercase{#1}}\xspace}}
\tsc{WGM}
\tsc{QE}

\begin{document}
\let\WriteBookmarks\relax
\def\floatpagepagefraction{1}
\def\textpagefraction{.001}

\shorttitle{Efficient Point Transformer with Dynamic Token Aggregating for LiDAR Point Cloud Segmentation}    

\shortauthors{D. Lu et al.}  

\title[mode = title]{Efficient Point Transformer with Dynamic Token Aggregating for LiDAR Point Cloud Segmentation}

\author[1]{Dening Lu}[orcid=0000-0003-0316-0299]

\author[2]{Jun Zhou}

\author[1]{Kyle (Yilin) Gao}


\author[1]{Linlin Xu}[orcid=0000-0002-6833-6462]

\author[3]{Jonathan Li}
\cormark[1]

\begin{abstract}
   Recently, LiDAR point cloud segmentation has made great progress due to the development of 3D Transformers. However, existing 3D Transformer methods usually are computationally expensive and inefficient due to their huge and redundant attention maps. They also tend to be slow due to requiring time-consuming point cloud sampling and grouping processes. To address these issues,  we propose an efficient point Trans\textbf{F}ormer with \textbf{D}ynamic \textbf{T}oken \textbf{A}ggregating (\textbf{DTA}-\textbf{F}ormer) for LiDAR point cloud segmentation. Firstly, we propose an efficient Learnable Token Sparsification (LTS) block, which considers both local and global semantic information for the adaptive selection of key tokens. Secondly, to achieve the feature aggregation for sparsified tokens, we present the first Dynamic Token Aggregating (DTA) block in the 3D Transformer paradigm, providing our model with strong aggregated features while preventing information loss. After that, a dual-attention Transformer-based Global Feature Enhancement (GFE) block is used to improve the representation capability of the model. Lastly, a novel Iterative Token Reconstruction (ITR) block is introduced for dense prediction whereby the semantic features of tokens and their semantic relationships are gradually optimized during iterative reconstruction. Based on ITR, we propose a novel W-net architecture, which is more suitable for Transformer-based feature learning than the common U-net design. Extensive experiments demonstrate the superiority of our method in LiDAR point cloud segmentation. It achieves SOTA performance with over 8$\times$ faster than prior point Transformers on the airborne MultiSpectral LiDAR (MS-LiDAR) dataset and aerial LiDAR dataset (DALES), and also surpasses prior efficient Transformers. The effectiveness of the proposed blocks has also been verified in an part segmentation dataset (ShapeNet) for clear visualization.
\end{abstract}


\maketitle
\section{Introduction}
\label{sec:introduction}
Thanks to advancements in sensor technology and algorithmic innovations, LiDAR point cloud segmentation  has been successfully applied to various fields such as 3D mapping, autonomous navigation, and city information modeling \citep{song2022training, abbasi2022lidar, cheng2022novel, wei2023dynamic}. Especially in the realm of remote sensing, point clouds prove indispensable by providing detailed and accurate information about the Earth's surface, aiding in tasks such as terrain mapping, environmental monitoring, and disaster response \citep{chen2022crackembed, lin2022semantic, chen2023adaptive}. Their ability to offer a comprehensive understanding of topography and land features underscores their significance in enhancing the capabilities of remote sensing technologies.

Recently, the Transformer has seen tremendous progress in point cloud processing and analysis because of its superior capability in long-range dependency modeling. Moreover, the inherent permutation invariance property of its self-attention mechanism aligns seamlessly with unordered point cloud data, which makes the Transformer highly suitable for point cloud processing. Point Transformer \citep{guo2021pct} and Point Cloud Transformer \citep{zhao2021point} explored applying Transformer to the entire point set and local regions respectively, and both of them achieved great success in point cloud segmentation. They demonstrated the great potential of Transformer in the field of point cloud processing. However, existing 3D Transformer methods usually suffer from high computational costs and low algorithmic efficiency, which are mainly caused by huge yet redundant attention maps tailored for all input points, as well as time-consuming point cloud sampling and grouping processes. Currently, most 3D Transformer methods \citep{zhao2021point, qiu2021geometric, lai2022stratified, zhang2022patchformer, robert2023efficient} apply hierarchical network structures, aiming to improve the model efficiency by gradually reducing input points. However, on the one hand, the commonly used Farthest Point Sampling (FPS) is not only time-consuming but also only focuses on the geometric properties while ignoring the semantic features of point clouds. This causes deep learning models to downplay certain fine-level object parts with significant semantic information. On the other hand, it is usually challenging for traditional point cloud grouping methods based on $k$NN ($k$-Nearest Neighborhood) to generate semantically homogeneous neighborhoods, thereby negatively impacting local feature aggregation. The aforementioned shortcomings are especially noticeable when handling extensive LiDAR datasets consisting of millions of points, leading to excessively high computational and memory expenses. Similar to superpixel in image processing \citep{achanta2012slic}, there are several works \citep{landrieu2018large, sun2023superpoint, robert2023efficient} proposed to explore combining superpoints with the Transformer for efficient point cloud processing. However, the static superpoint clustering strategy cannot adaptively serve the semantic features extracted at different stages of the network. Additionally, the max/average pooling operation used in superpoint generation tends to cause information loss.

Regarding these issues, we propose a novel Transformer-based point cloud representation framework, named DTA-Former which is the first work to introduce the dynamic token selection, aggregation, and reconstruction methods to 3D Transformers. Firstly, a Learnable Token Sparsification (LTS) block is proposed for the adaptive and efficient selection of key tokens. Compared with FPS, it not only considers both local and global semantic information for decision-making but also significantly reduces the computational cost. Secondly, instead of local feature aggregation by the max/average pooling operation, a Weighted Cross-Attention (WCA)-based Dynamic Token Aggregating (DTA) block is designed for efficient token aggregation, aggregating semantic information for each sparsified token at a global scale. It provides our model with strong aggregated features while mitigating information loss. Finally, we propose a new W-net architecture equipped with additional Iterative Token Reconstruction (ITR) blocks for point cloud segmentation. The ITR block is placed at the end of \textit{each stage} of the network, mapping the sparsified tokens back into the original token set by utilizing their semantic relationships. Extensive experiments on various tasks demonstrate the superiority of the proposed DTA-Former. Towards the airborne MS-LiDAR \citep{zhao2021airborne} and aerial DALES \citep{varney2020dales} datasets, our model obtains excellent results with 88.3$\%$ average $F_{1}$ score and 78.5$\%$ mIoU respectively, outperforming prior SOTA methods.

The main contributions of our work can be summarized as follows:
\begin{itemize}
    \item We propose a novel Learnable Token Sparsification (LTS) block, which considers both local and global semantic information for efficient and adaptive selection of key points. It surpasses commonly used sampling approaches such as FPS in terms of both accuracy and efficiency. 
    \item We present the first Dynamic Token Aggregating (DTA) block in the 3D Transformer paradigm. It is realized by using a novel Weighted Cross-Attention (WCA) mechanism for semantic feature aggregation, enlarging the effective receptive field while mitigating information loss.
    \item We propose a new Iterative Token Reconstruction (ITR) block for the dense prediction of point clouds. It maps the fine-aggregated tokens into the original token set through the cross-attention map generated by WCA (named WCA-map). The accurate token relationships captured by the WCA-map ensure precise token reconstruction.
    \item We present DTA-Former incorporating the aforementioned blocks for efficient point cloud learning and representation. A novel W-net architecture is designed for point cloud dense prediction. Extensive experiments on the airborne MS-LiDAR and aerial DALES semantic segmentation datasets demonstrate the superiority of our method in LiDAR point cloud segmentation, in terms of both accuracy and efficiency, surpassing the prior SOTA methods.
\end{itemize}

The remainder of our paper is organized as follows. Section \ref{sec:relatedwork} reviews existing 3D Transformer methods and summarizes the limitations. Section \ref{sec:method} shows the details of DTA-Former. Section \ref{sec:Experiments} presents and discusses the experimental results. Section \ref{sec:conclusion} concludes the paper.

\section{Related Work}
\label{sec:relatedwork}

\subsection{Point Cloud Transformers}
Point cloud Transformers have emerged as a powerful paradigm for 3D data processing, demonstrating notable success in tasks such as object recognition, scene understanding, and segmentation. Their permutation invariance and ability to capture complex spatial relationships have contributed to advancements in point cloud segmentation. Existing point cloud Transformers can be broadly categorized into two main groups: global Transformer-based methods and local Transformer-based methods. Here, we review existing approaches in both categories and summarize the limitations. 

The global Transformer methods \citep{guo2021pct, hui2021pyramid, kaul2022convolutional, zhang2022introducing, zhang2022patchformer, han20223crossnet, sun2023superpoint, robert2023efficient, 10273676, li2023gl} are designed to capture long-range dependencies across the entire point cloud, which is the most straightforward approach in 3D Transformer designing. Point Cloud Transformer (PCT) \citep{guo2021pct} is the first work to introduce Transformers into global feature learning of point clouds. The encoder of PCT fed neighborhood-embedded features into a series of stacked Transformer blocks, followed by using MA-Pool (Max-Pool $+$ Average-Pool) to extract global features. Its decoder consists of several MultiLayer Perceptrons (MLPs) separated by residual connections for point cloud segmentation, following PointNet \citep{qi2017pointnet}. MPT$+$ \citep{zhang2022introducing} improved the self-attention mechanism in the Transformer block for land cover classification in remote sensing. The proposed BiasFormer adds a positional bias to encode the geometric relationship, enhancing the spatial modeling ability of the Transformer. The main drawback of global Transformer methods is the high computational and memory costs, which is caused by its $\mathcal{O}(N^{2}D)$ complexity, where $N$ is the number of input points, and $D$ is the feature dimension. The runtime of global Transformer methods grows quadratically as the number of input points grows \citep{liu2023flatformer}. Therefore, it is challenging for global Transformer methods to process large-scale scenes for remote sensing.



The local Transformer methods \citep{zhao2021point, lai2022stratified, gao2022lft, liu2023flatformer} are designed to extract local information on a group of subsets of the target point cloud, rather than the entire object. Therefore, their computational complexity is $\mathcal{O}(Nk^{2}D)$, where $k$ is the number of points in the local subset. Due to the unordered nature of point clouds, local Transformers can be further divided into two categories: the neighborhood-based Transformer and window-based Transformer, according to their partition strategies for subset generation. Point Transformer \citep{zhao2021point} is a representative work of neighborhood-based local Transformers. With a U-net architecture, it collected the neighborhood for each sampling point in the encoder using $k$NN search, where the Transformer block is used for local feature extraction. Its decoder is symmetric to the encoder, with trilinear interpolation \citep{qi2017pointnet++} for point cloud upsampling. Stratified Transformer \citep{lai2022stratified}, as a window-based local Transformer, took 3D voxels as input and applied Transformers in predefined local windows, following Swin Transformer \citep{liu2021swin}. To capture the global information and establish connections between different windows, it presented a multi-scale $key$ sampling strategy, enlarging the effective receptive field for each $query$ point. FlatFormer \citep{liu2023flatformer} also used Transformer blocks to extract window-based local features. It flattened the point cloud with window-based sorting and designed a window shift strategy to achieve global feature learning. However, such indirect global feature learning methods limit the performance of point cloud Transformers in long-range context dependency modeling. Additionally, local Transformers usually utilize pooling operations for local feature aggregation, such as max pooling and average pooling, which tend to cause local information loss. 

\subsection{Efficient Transformers}
Efficient Transformers are a family of Transformer models designed to improve the efficiency and scalability of traditional Transformer architectures while maintaining or even surpassing their performance on various tasks. In the field of Natural Language Processing (NLP), the birthplace of the Transformer algorithm, substantial and insightful works \citep{DaiYYCLS19, AinslieOACFPRSW20, MartinsMM22, TayDBM23, NawrotCLP23} have been done in the realm of efficient Transformers. The recent dynamic-pooling Transformer \citep{NawrotCLP23} reduced sequence length by predicting segment boundaries for input texts in an autoregressive fashion. By this means, the input tokens could be segmented into a series of token subsets. It proposed a dynamic token pooling approach, thereby enhancing network efficiency and reducing algorithmic complexity. However, due to the unordered nature of 3D point clouds and the significantly larger number of input tokens than NLP data, it is not feasible to directly apply the boundary prediction algorithm of \citep{NawrotCLP23} to 3D token clustering. Our network framework utilizes a novel hierarchical structure equipped with LTS, DTA, and ITR blocks for strong point cloud feature learning, which is tailored to the representation of 3D point clouds. 

Recently, there also have been many efficient point cloud Transformers \citep{hui2021pyramid, zhang2022patchformer, park2022fast, sun2023superpoint, robert2023efficient, wang2023dsvt, liu2023flatformer} proposed to reduce computation and memory costs and improve processing efficiency. The hierarchical processing framework is one of the most common techniques in point cloud processing, especially for large-scale LiDAR point clouds. PPT-Net \citep{hui2021pyramid} proposed a hierarchical encoder-decoder network to reduce the number of points gradually. Instead of using a pure Transformer architecture, it combined graph convolution-based \citep{wang2019dynamic} local feature embedding and Transformer-based global feature learning, which not only enhances long-term dependencies among points but also reduces the computational cost. To reduce the number of input tokens and improve the models' ability to local feature extraction, PatchFromer \citep{zhang2022patchformer}, SPFormer \citep{sun2023superpoint}, and SPT \citep{robert2023efficient} all designed superpoint-based local feature aggregation methods, based on positional and radiometric information from raw point clouds. They are able to generate geometrically-homogeneous point clusters for further local feature extraction, and only need to be calculated once, as a pre-processing step. However, the static point clustering strategy they used can not adaptively serve the semantic features extracted at different stages of the network, limiting their performance. 

To address the aforementioned drawbacks, we propose a novel Transformer-based point cloud representation framework, DTA-Former, designing the first dynamic token selecting, aggregating, and reconstruction method for point cloud processing. Firstly, different from traditional downsampling approaches, we introduce a new LTS block, selecting key tokens adaptively by considering both global and local semantic features. It is able to greatly reduce the computational cost. Secondly, instead of commonly used feature pooling operations, a novel WCA-based DTA block is designed for efficient token aggregation, enlarging the effective receptive field while mitigating information loss. Finally, instead of using trilinear interpolation for point cloud upsampling, we design an efficient ITR block for dense prediction, utilizing semantic relationships between tokens to ensure a simple yet highly accurate reconstruction operation.

\begin{figure*}[htbp]
  \centering
  \includegraphics[width=\linewidth]{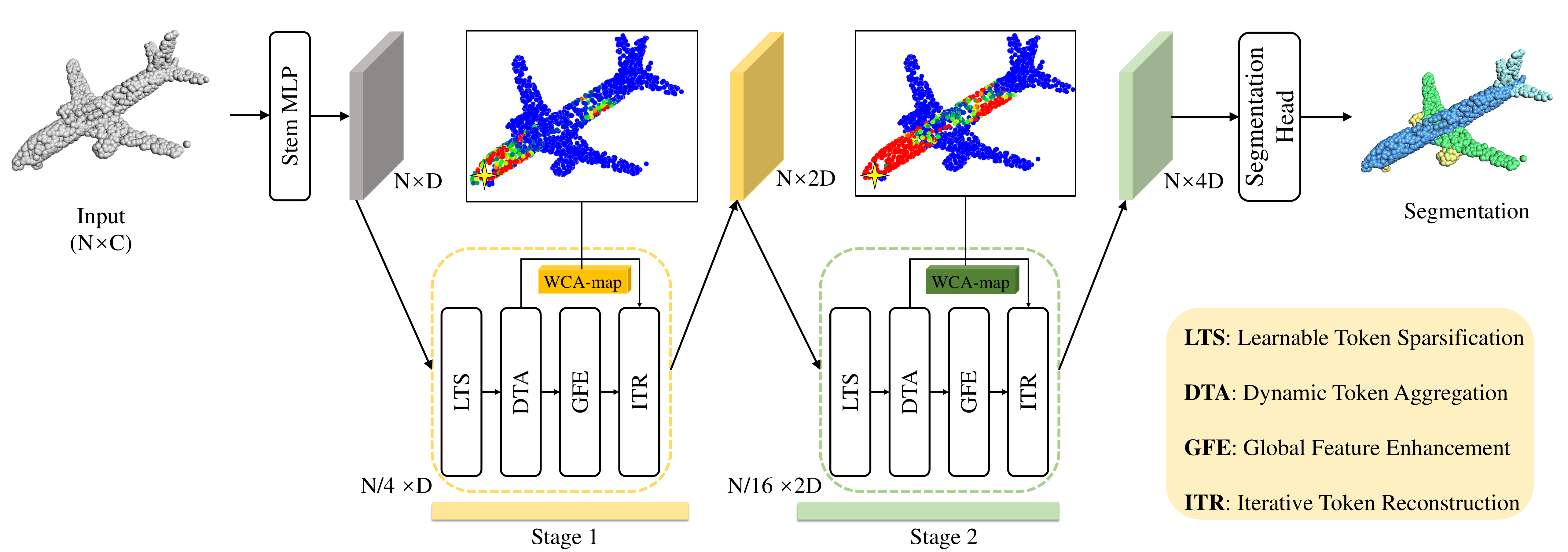}
  \caption{Architecture of DTA-Former. A novel W-net architecture is proposed, with two stages containing LTS, DTA, GFE, and ITR blocks. \textit{To clearly visualize the semantic homogeneous clustering results, the airplane model in ShapeNet \citep{yi2016scalable} is taken as an example to illustrate the details of the method.} The two WCA-maps show strong correlations between attention weights and point semantics, where the stronger the correlation, the redder the point. Query points are indicated by yellow stars.
  \label{fig:overview}}
\end{figure*}

\section{Method}
\label{sec:method}
This section details the design of DTA-Former. We first present the overall framework, then introduce its key components: LTS, DTA, GFE, and ITR blocks.

\subsection{Overview}
The overall pipeline of DTA-Former is shown in Fig. \ref{fig:overview}. To clearly visualize the semantic homogeneous clustering results, we use the CAD-like model in ShapeNet \citep{yi2016scalable} to demonstrate the overall process.
Instead of using the common U-net-style encoder-decoder architecture, we design a novel W-net-style architecture for dense prediction. Compared with the U-net design \citep{qi2017pointnet++}, our network is more suitable for Transformer-based feature learning.
The input point cloud is firstly fed into a stem MLP block \citep{qian2022pointnext} for higher-dimension token generation. Secondly, it is fed into several stages hierarchically for token sparsification, aggregating, and global feature learning. Specifically, each stage consists of four blocks: a LTS block, a DTA block, a GFE block, and an ITR block. The first three blocks are designed for efficient downsampling and fully exploring long-range dependencies among tokens. The ITR block is designed to efficiently maps the sparsified tokens back into the original token space, gradually optimizing the semantic features of tokens and their semantic relationships during iterative reconstruction. Lastly, an MLP head layer consisting of a max-pooling layer, two fully connected layers, batch normalization, and RELU activation is placed at the end of each stage, to generate the final prediction for each point. The number of output tokens from each stage is the same as the raw input tokens, allowing us to utilize a multi-loss function to optimize each stage effectively. The details of the LTS, DTA, GFE, and ITR block are elaborated in the following subsection, as well as W-net architecture.

\begin{figure*}[htbp]
  \centering
  \includegraphics[width=\linewidth]{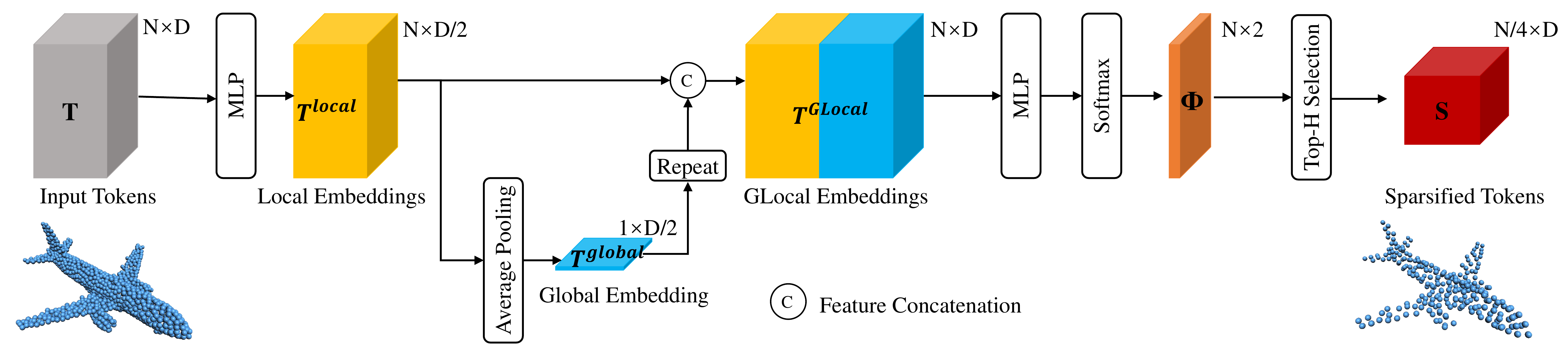}
  \caption{Illustration of LTS results on Stage-1. Utilizing both local and global semantic information of tokens, the token sparsification process is dynamically and efficiently updated as the computation proceeds.
  \label{fig:lts}}
\end{figure*}

\begin{figure}[b]
  \centering
  \includegraphics[width=\linewidth]{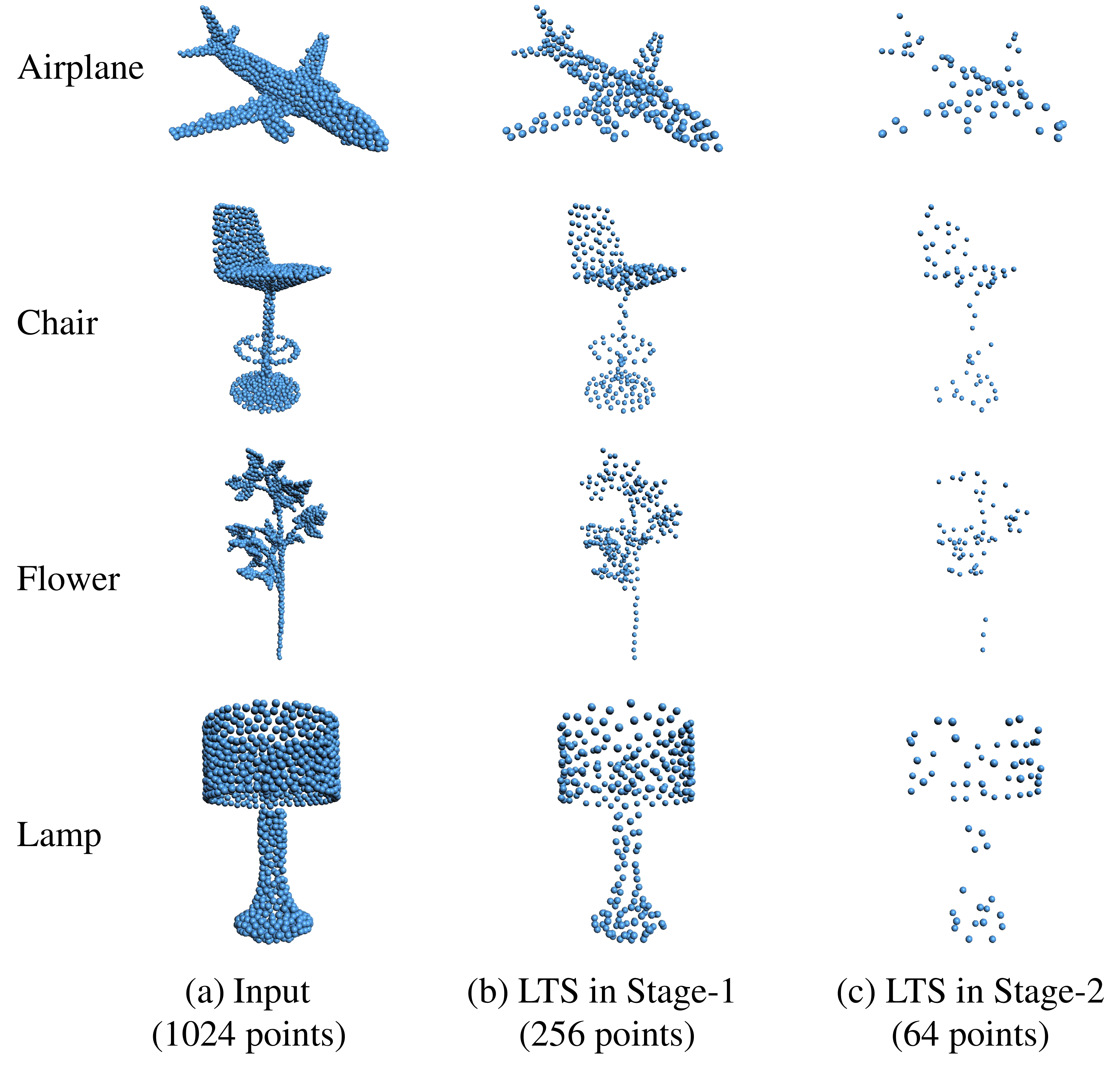}
  \caption{Visualization of sparsified tokens from LTS blocks for different object categories. As observed, LTS works well in keeping the structural skeleton of objects, even in scenarios with very sparse sampling tokens.
  \label{fig:lts_result}}
\end{figure}

\subsection{Learnable Token Sparsification (LTS)}
\label{subsec:LTS}
LTS drops redundant tokens dynamically according to learnable decision scores. As shown in Fig. \ref{fig:lts}, given the input tokens $T = \left \{t_{i} \right \}_{i=1}^{N} \in R^{N \times D}$, where $N$ is the number of input tokens, and $D$ is the feature dimension, we first feed them into an MLP, getting the local embeddings:
\begin{equation}
T^{local}= MLP(T) \in R^{N \times D/2},
\end{equation}
where $T^{local}$ is the local embedding of $T$. Furthermore, we compute a global feature for input tokens by using average pooling, allowing the LTS block to analyze the importance of each token from a global perspective:
\begin{equation}
T^{global}= Ave(MLP(T)) \in R^{1 \times D/2},
\end{equation}
where $T^{global}$ is the global embedding of $T$. We then concatenate ${T}^{local}$ and $T^{global}$ to get the GLocal (global $+$ local) embedding for $T$, which contains both single token information and context information from the whole set:
\begin{equation}
T^{GLocal}= \left [ T^{local}, T^{global}  \right ] \in R^{N \times D}.
\end{equation}
Given GLocal embedding set $T^{GLocal}$, we compute the decision score set $\Phi = \left \{\varphi _{i} \right \}_{i=1}^{N} \in R^{N \times 2}$ for all tokens in $T$:
\begin{equation}
\Phi = Softmax(MLP(T^{GLocal})) \in R^{N \times 2}.
\end{equation}
$\Phi_{i}$ has two feature channels representing keeping ($\varphi _{i,0}$) and dropping ($\varphi _{i,1}$) probabilities, respectively. Finally, Gumbel-Softmax \citep{jang2016categorical} is employed to select the top-$H$ tokens with the highest keeping probabilities, ensuring a differentiable process. After the token sparsification process, we obtain a subset of $T$, which is denoted as $S = \left \{s_{i} \right \}_{i=1}^{H} \in R^{H \times D}$. Thanks to the learnable decision scores, the token sparsification process is dynamically and efficiently updated as the computation proceeds. 
The visualization results of sparsified tokens on different object categories are shown in Fig. \ref{fig:lts_result}. 

\begin{figure*}[htbp]
  \centering
  \includegraphics[width=\linewidth]{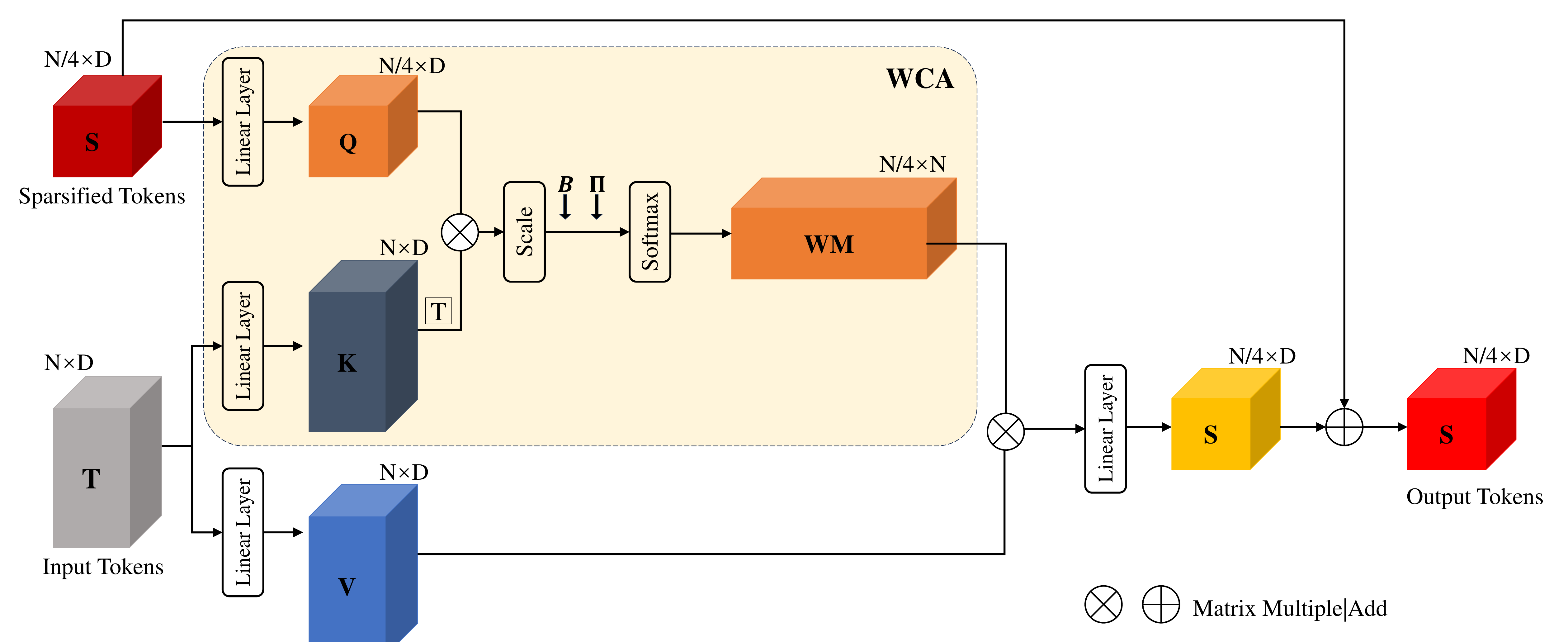}
  \caption{Illustration of the DTA block on Stage-1. It exploits the long-range context information better for feature aggregating without information loss. It is more efficient than local grouping and pooling operations due to avoiding neighborhood construction.
  \label{fig:DTA}}
\end{figure*}

\subsection{Dynamic Token Aggregating (DTA)}
\label{subsec:DTA}
The commonly used pooling approaches like max and average pooling which operate in local neighborhoods for feature aggregation. This limits the receptive field and tends to cause information loss. Thanks to the great capability of long-range dependency modeling of the self-attention mechanism, we propose a DTA block by a new Weighted Cross-Attention (WCA) mechanism. It fully exploits the semantic relationships between $S$ and $T$ and achieves long-range feature aggregating adaptively for each sparsified token $s_{i}$, enlarging the effective receptive field while mitigating information loss. As shown in Fig. \ref{fig:DTA}, given the sparsified token set $S$, we first generate $Query$ matrix based on $S$, and $Key$, $Value$ matrices based on $T$:
\begin{equation}
\begin{aligned}
Q =  S W_{Q},\\
K =  T W_{K},\\
V =  T W_{V},\\
\label{eq:wca}
\end{aligned}
\end{equation}
where $Q$, $K$, $V$ represent $Query$, $Key$, $Value$ matirces, $W_{Q}$, $W_{K}$, $W_{V}$ are learnable weight matrices. Secondly, the WCA-map $WM$ is calculated as:
\begin{equation}
WM = Softmax(\Pi \odot (\frac{QK^{T}}{\sqrt{D}}+B)) \in R^{H \times N},
\label{eq:wm}
\end{equation}
where $B$ is a learnable position encoding matrix defined by \citep{zhao2021point}, $\odot$ represents the Hadamard operation, and $\Pi = \left \{\varphi _{i, 0} \right \}_{i=1}^{N} \in R^{N \times 1}$ is a vector of decision scores in Section \ref{subsec:LTS}, whose components are used as weights for input tokens. 
In detail, since the shape of $\Pi$ is $N \times 1$ which is inconsistent with $QK^{T}$, we first reshape $\Pi$ to $1 \times N$, followed by expanding it to $H \times N$ by row repetition.
$\Pi$ essentially represents the importance of each point to network decision-making. The combination of $\Pi$ and the attention map allows the model to focus more on key tokens with high scores, which is beneficial to enhance the representation ability of the network. The ablation study in Section \ref{subsec:ablation} also demonstrates the superiority of WCA. Finally, $WM$ is taken as a weight matrix for global token aggregating, and multiplied by $V$ to generate the final pooled sparsified token set $S$ with the size of $H \times D $. Compared with local grouping and pooling operations, DTA exploits the long-range context information better for feature aggregating, while being more efficient due to avoiding neighborhood construction.

\subsection{Global Feature Enhancement (GFE)}
\label{subsec:GFE}
We further enhance the global feature learning of the pooled tokens by a dual-attention Transformer block, thanks to its remarkable ability of long-range context dependency modeling. It fully explores the relationships between individual points (point-wise self-attention) as well as the associations among different channels (channel-wise self-attention), resulting in better featurization and representation of the model than vanilla point-wise or channel-wise Transformers. The process of point-wise and channel-wise self-attention is formulated as follows:
\begin{equation}
\begin{aligned}
&F_{P} = softmax(\frac{\left ( SW_{QE} \right )\left ( SW_{KE} \right )^{T}}{\sqrt{D}} + \grave{B} ) \left ( SW_{VE} \right ), \\
&F_{C} = softmax(\frac{\left ( SW_{KE} \right )^{T} \left ( SW_{QE} \right )}{\sqrt{D}}) \left ( SW_{VE} \right )^{T},
\end{aligned}
\end{equation}
where $F_{P}$ is the enhanced global features of $S$ from the point-wise self-attention branch, while $F_{C}$ is the enhanced global features from the channel-wise self-attention branch. $W_{QE}$, $W_{KE}$, $W_{VE}$ are learnable weight matrices, which are similar to Eq. \ref{eq:wca}. $\grave{B}$ is a learnable position encoding matrix defined by \citep{zhao2021point}. We finally combine $F_{P}$ and $F_{C}$ by the element-wise addition, followed by a residual connection and a linear projection to increase the feature dimension for subsequent calculations:
\begin{equation}
S = LBR\left (S + MLP\left (F_{P} + F_{C} \right )\right ),
\end{equation}
where $LBR$ denotes the combination of $Linear$, $BatchNorm$, and $ReLU$. As such, we achieve the dual Transformer-based global feature enhancement for $S$. Ablation experiments in Section \ref{subsec:ablation} investigated the effectiveness of the dual-attention mechanism, which demonstrates its superiority compared with vanilla point-wise or channel-wise mechanisms. 

\begin{figure*}[htbp]
  \centering
  \includegraphics[width=0.95\linewidth]{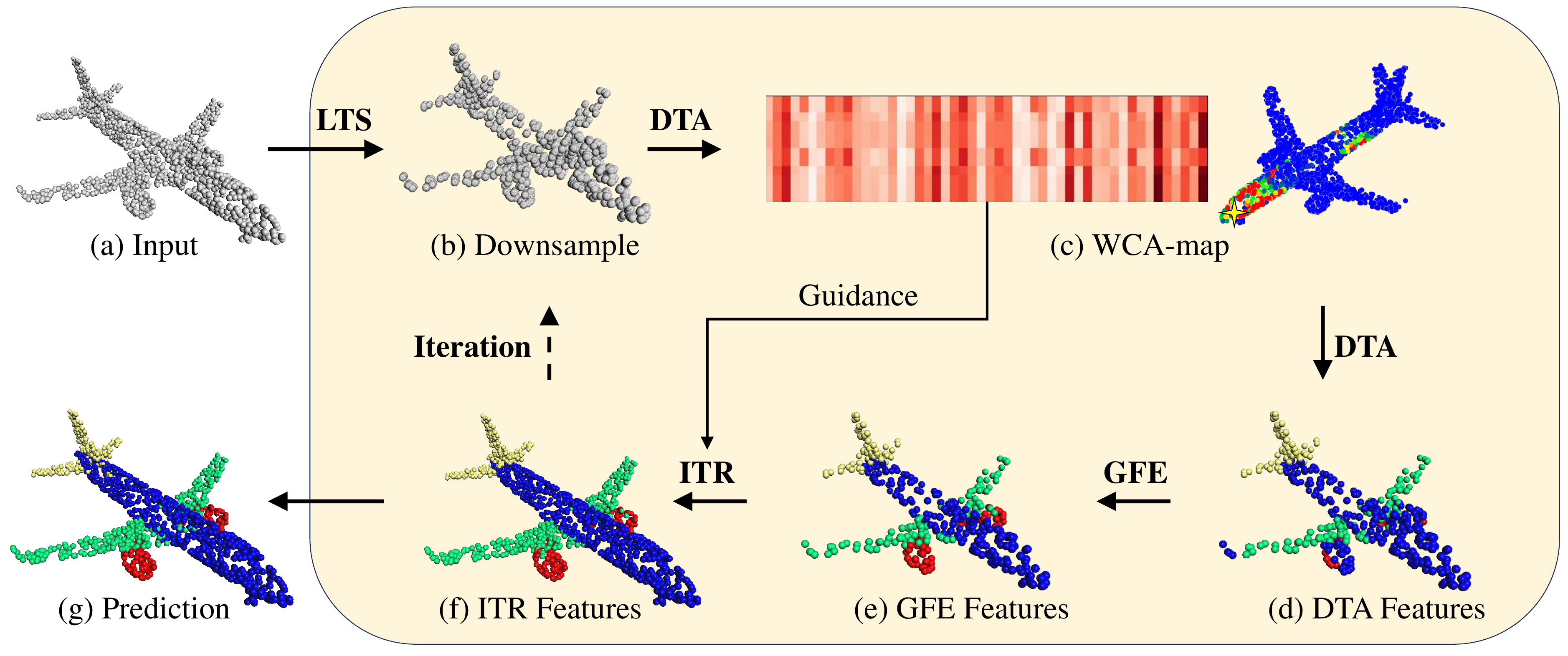}
  \caption{Brief illustration of the segmentation network.
  \label{fig:ITR_pipeline}}
\end{figure*}

\begin{table*}[htbp]\color{black}
 \caption{Confusion matrix (\%) of DTA-Former on the airborne MS-LiDAR dataset. The numbers in the last three rows represent the precision, recall, and $F_{1}$ score for each class. \label{tab:semseg_confusion}
 }
 \centering
 \setlength{\tabcolsep}{15pt}
 \renewcommand{\arraystretch}{1.2}
 \begin{tabular}{llllllll}
  \hline
  \multicolumn{2}{l}{\multirow{2}{*}{\textbf{Categories}}}  &\multicolumn{6}{l}{\textbf{True Label}} \\
   \cline{3-8}
    & &Road & Building & Grass  & Tree & Soil & Powerline  \\
  \hline
  \multirow{6}{*}{\rotatebox{90}{\textbf{Prediction Label}}} &\multicolumn{1}{l}{Road} & 82.5  & 1.5   & 0.0  &0.0 & 5.3 & 0.0   \\
  \cline{2-8}
  &\multicolumn{1}{l}{Building} & 14.1  & 95.7   & 0.7  & 0.3 &38.6 & 0.0   \\
  \cline{2-8}
  &\multicolumn{1}{l}{Grass} & 0.4  & 1.2   & 98.8  & 2.0 &0.1 & 13.4   \\
  \cline{2-8}
  &\multicolumn{1}{l}{Tree} & 0.1  & 0.2   & 0.4  & 97.7 &0.2 & 1.2   \\
  \cline{2-8}
  &\multicolumn{1}{l}{Soil} & 2.8  & 1.2   & 0.0  & 0.0 &56.0 & 0.0   \\
  \cline{2-8}
  &\multicolumn{1}{l}{Powerline} & 0.0  & 0.0   & 0.0  & 0.0 &0.0 & 85.2   \\
 \hline
 \hline
 \multicolumn{2}{l}{Precision} & 93.1 & 89.1   & 99.0  & 94.1 &77.4 &94.3   \\
  \hline
 \multicolumn{2}{l}{Recall} & 82.5  & 95.9   & 98.8  & 97.7 & 55.9 & 86.3  \\
  \hline
 \multicolumn{2}{l}{$F_{1}$} & 87.5  & 92.4  & 98.9  & 95.9 &64.9 &90.1  \\
  \hline
 \end{tabular}
\end{table*}

\subsection{Iterative Token Reconstruction (ITR)}
\label{subsec:upsampling}
Given the enhanced token set $S$, we map it back into the original token set $T$ by the WCA-map $WM$ in Eq. \ref{eq:wm}, followed by a residual connection:
\begin{equation}
T = Softmax(WM^{T})S + T \in R^{N \times D}.
\end{equation}
As shown in Fig. \ref{fig:overview}, the ITR block is placed in \textit{every stage} in the segmentation network, which leads to the W-net architecture. As such, we can reconstruct input tokens dynamically according to the WCA-map stored in DTA, and gradually optimize the semantic features of tokens and their semantic relationships during iterative reconstruction. To show the relationship between the WCA-map and the ITR block clearly, we provide a brief illustration of the segmentation network in Fig. \ref{fig:ITR_pipeline}. We also tried to apply the ITR block to the U-net architecture for segmentation (Section \ref{subsec:ablation}). However, the performance of the U-net drops significantly compared with the W-net. This is because the symmetric structure of the U-net hinders WCA-maps stored in its encoder from accurately describing the feature relationships within the sparsified token set that has undergone deep feature learning in the decoder. In contrast, our segmentation network allows for the prompt feedback of WCA-maps to sparsified tokens for reconstruction.

\section{Experiments}
\label{sec:Experiments}
In this section, we first present the implementation details of DTA-Former, including hardware configuration, training strategy, and hyperparameter settings. Secondly, we present the performance of DTA-Former on the LiDAR remote sensing semantic segmentation task (Airborne MS-LiDAR dataset  \citep{zhao2021airborne} and DALES \citep{varney2020dales}), comparing it to SOTA methods. We also tested DTA-Former on the part segmentation dataset (ShapeNet \citep{yi2016scalable}) to verify the effectiveness of semantic homogeneity clustering in the method.
Lastly, we conducted ablation studies to verify the effectiveness of key components in DTA-Former, as well as explored the sensitivities of the model to input data.

\subsection{Implementation Details}
DTA-Former was implemented with PyTorch and runs on two NVIDIA GeForce RTX 3090 GPUs. It was trained with the SGD Optimizer, with a momentum of $0.9$ and weight decay of $0.0001$. The initial learning rate was set to $0.1$, with a cosine annealing schedule to adjust the learning rate at every epoch. The network was trained for $200$ epochs with a batch size of $8$ for the segmentation task.


\begin{figure*}[htbp]
  \centering
  \includegraphics[width=0.9\linewidth]{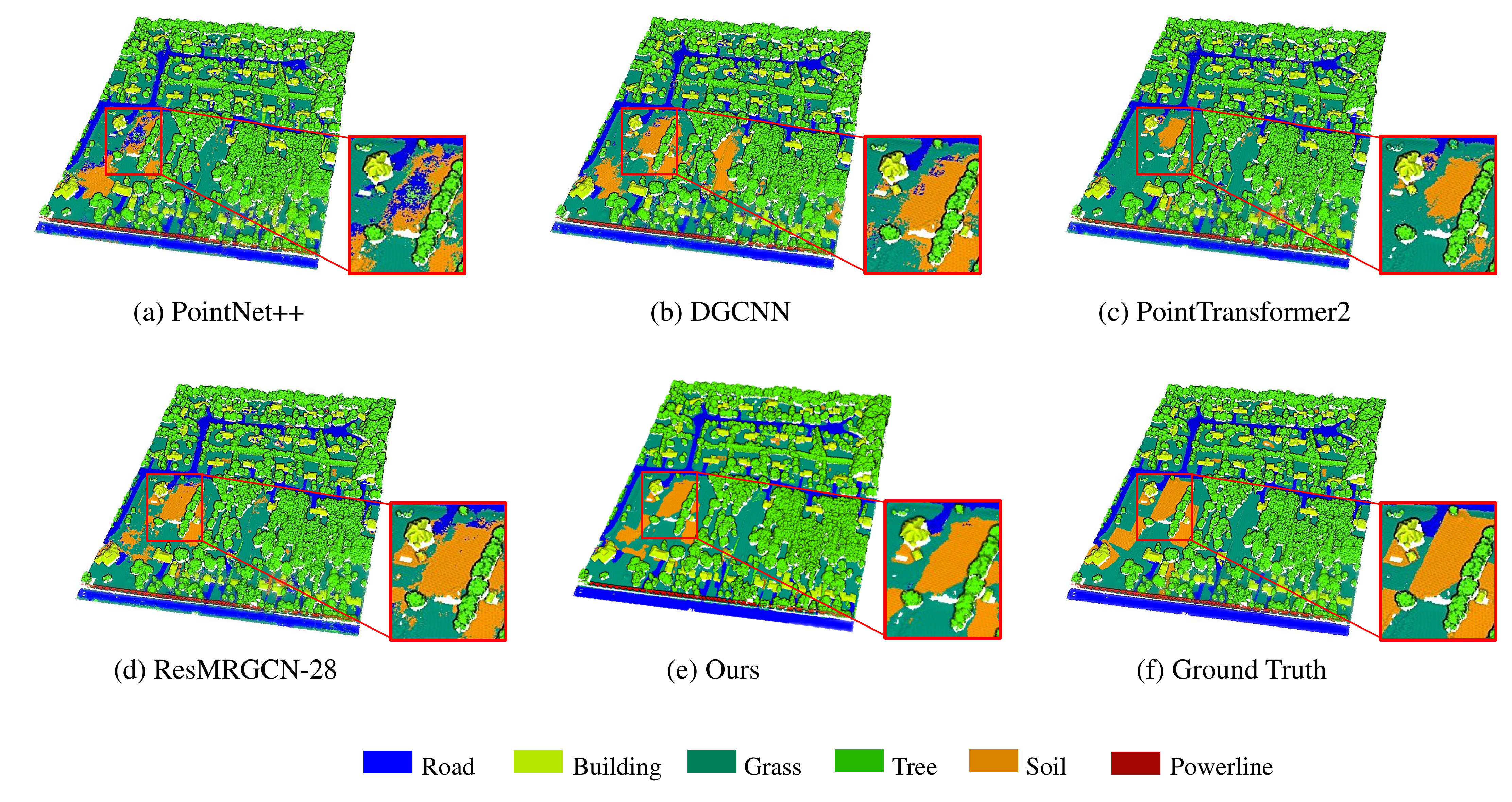}
  \caption{Comparison results from different methods on the testing area-11 in the airborne MS-LiDAR dataset.
  \label{fig:rs_comparison_11}}
\end{figure*}

\begin{table*}[htbp]
 \caption{Quantitative comparison ($\%$) of DTP-Former on the airborne MS-LiDAR dataset. The highest evaluation score is shown in bold type. \label{tab:rs_comparison}
 }
 \centering
 \setlength{\tabcolsep}{8pt}
 \renewcommand{\arraystretch}{1.0}
 \begin{tabular}{l|l|l|l|l}
  \hline
   {Methods}  & Average $F_{1}$ score & mIoU & OA & Latency (ms)  \\
  \hline
  {PointNet++} \citep{qi2017pointnet++}   & 72.1 & 58.6 & 90.1   & 322.6 \\
  {DGCNN} \citep{wang2019dynamic}  & 71.6 & 51.0 & 91.4   & 86.2   \\
  {RSCNN} \citep{liu2019relation} & 73.9 & 56.1 & 91.0  & 158.7  \\
  {GACNet} \citep{wang2019graph}   & 67.7 & 51.0 & 90.0  & 277.8 \\
 
  {AGConv} \citep{zhou2021adaptive} & 76.9 & 71.2 & 93.3 & 312.5  \\ 
  {SE-PointNet++} \citep{jing2021multispectral}   & 75.9 & 60.2 & 91.2  & - \\
  {FR-GCNet} \citep{zhao2021airborne} & 78.6 & 65.8 & 93.6 & - \\
  {PointTransformer} \citep{zhao2021point} & 80.5 & 73.6 & 93.1 & 285.7 \\
  {PPT-Net} \citep{hui2021pyramid}  & 80.1 & 73.6 & 92.7 & 43.3 \\
  {PatchFormer} \citep{zhang2022patchformer}   & 82.4 & 77.8 & 93.1 & 62.9  \\
  {Xiao et al.} \citep{xiao2022multispectral}   & 83.3 & 79.3 & 94.0  & - \\
  {ResMRGCN-28} \citep{9408381}   & 81.1 & 74.0 & 93.3  & 45.7\\
  {GCNAS} \citep{zhang2023semantic}   & 88.1 & \textbf{82.3} & 95.2 & - \\
  {PTv3} \citep{PTv3}   &85.5 & 77.1 &93.4 & 40.6\\
  {DCTNet} \citep{lu2024dynamic}   & 86.0 & 80.2 & 95.0 & 53.3 \\
  
    \hline
  {Ours}     &\textbf{88.3} & 81.4 & \textbf{95.4} & \textbf{32.8}\\

   \hline
 \end{tabular}
\end{table*}

\begin{figure*}[htbp]
  \centering
  \includegraphics[width=0.9\linewidth]{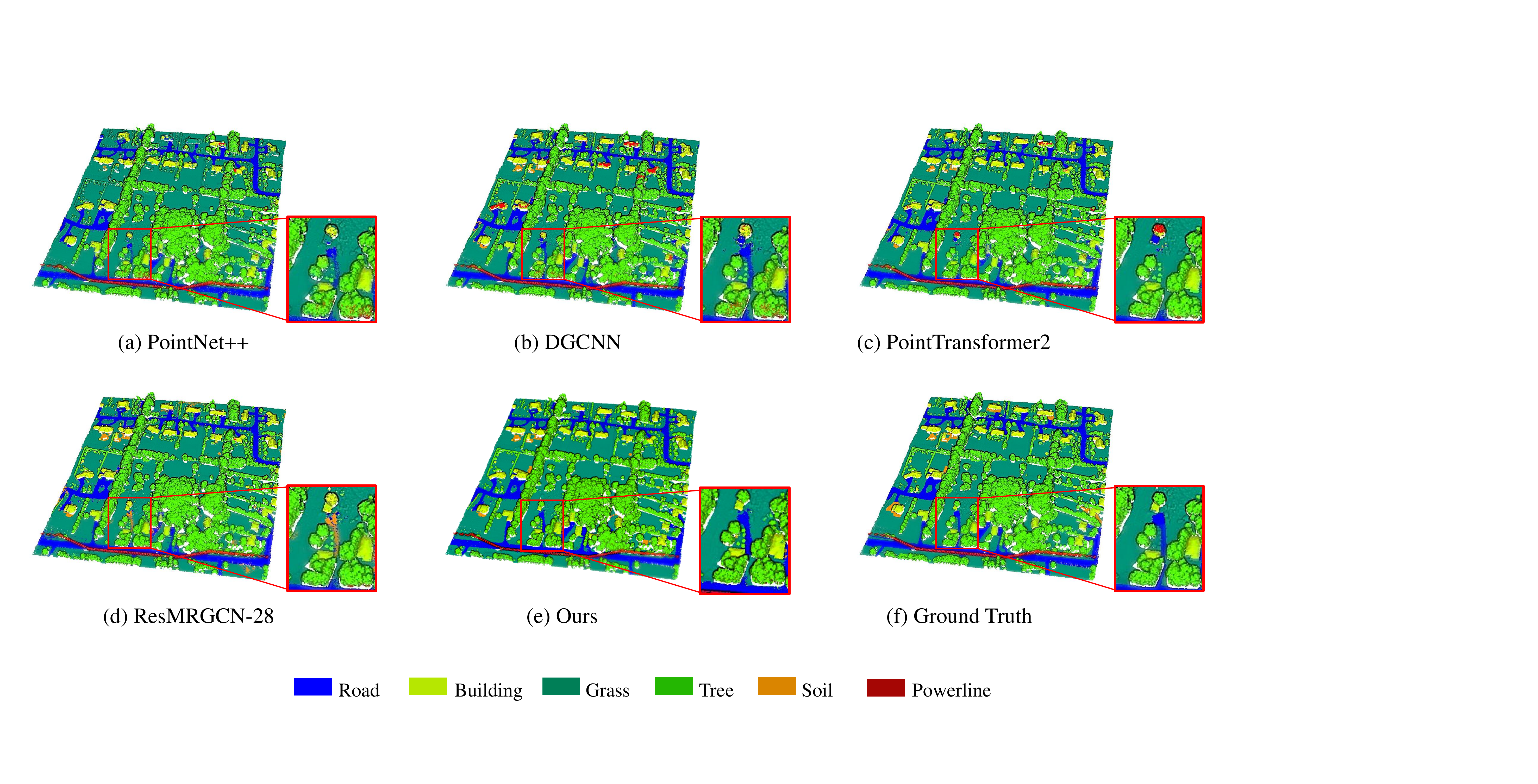}
  \caption{Comparison results from different methods on the testing area-12 in the airborne MS-LiDAR dataset.
  \label{fig:rs_comparison_12}}
\end{figure*}

\begin{figure*}[htbp]
  \centering
  \includegraphics[width=0.9\linewidth]{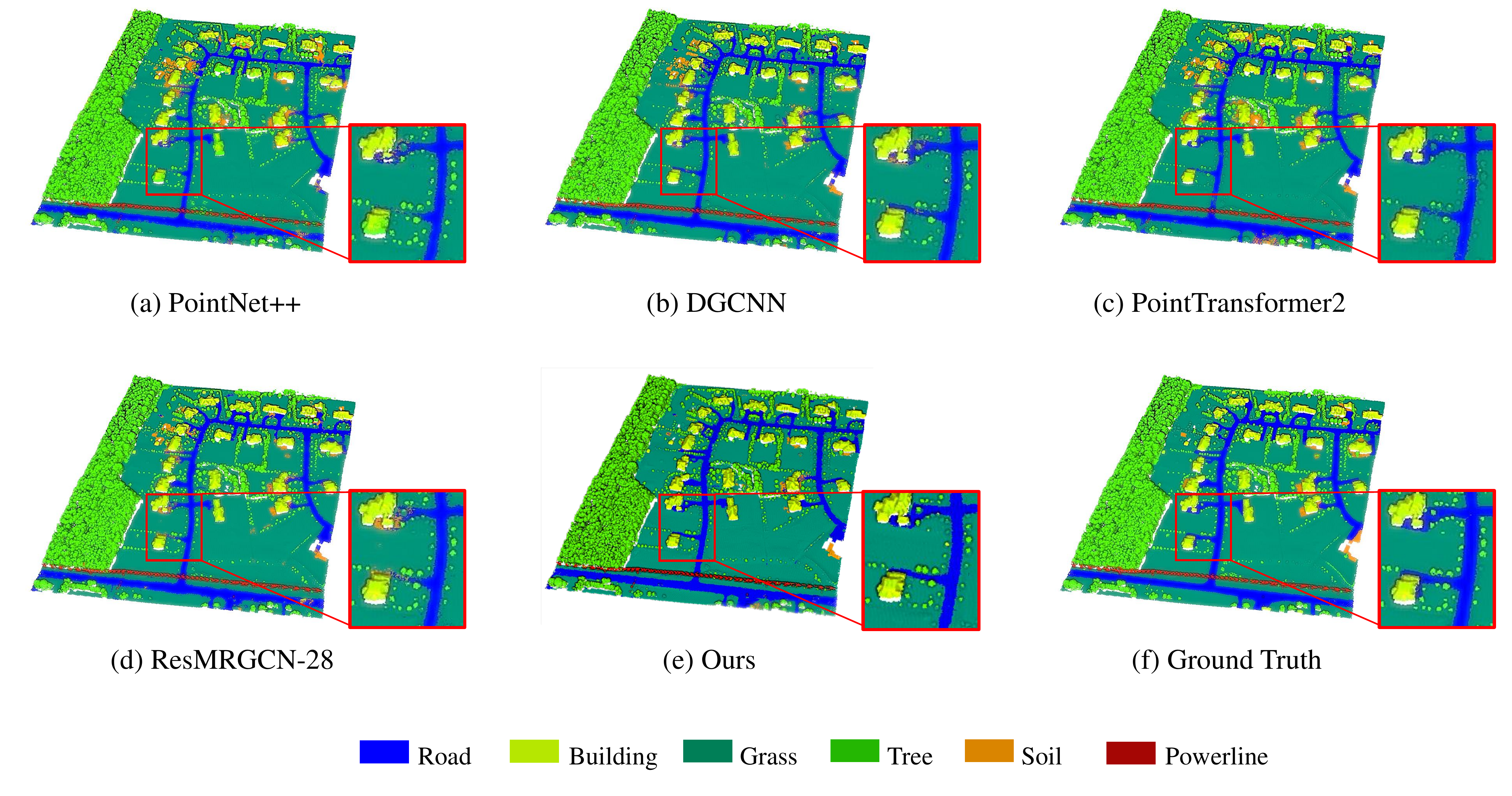}
  \caption{Comparison results from different methods on the testing area-13 in the airborne MS-LiDAR dataset.
  \label{fig:rs_comparison_13}}
\end{figure*}

\begin{figure*}[htbp]
  \centering
  \includegraphics[width=0.9\linewidth]{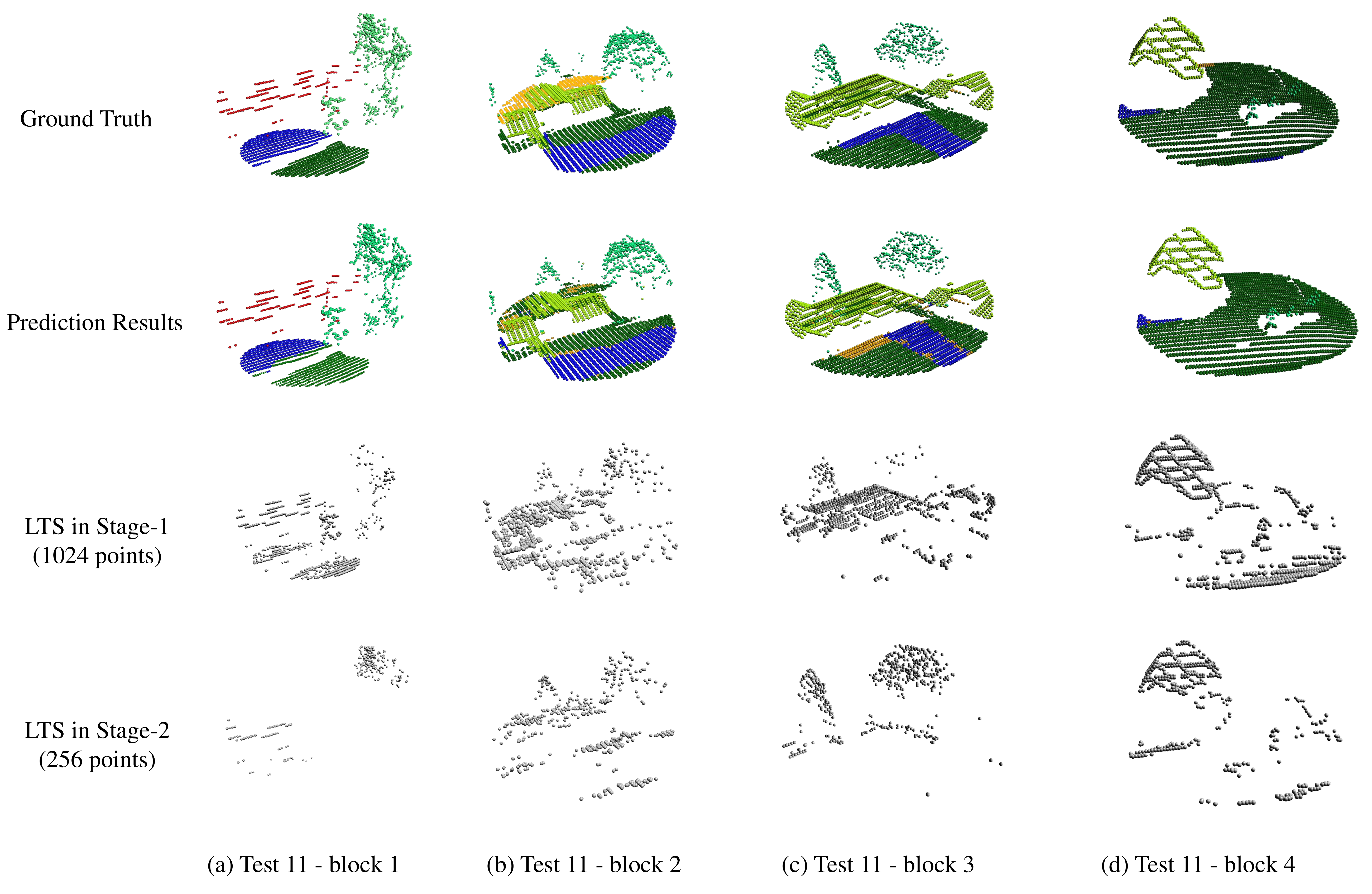}
  \caption{Visualization results of samples from the testing area-11. LTS results in different network stages are also shown. They confirm the excellent performance of DTA-Former in MS-LiDAR data segmentation and illustrate similar geometric characteristics of the soil and roads, which is the key reason why misclassified points are mainly located in soil areas.
  \label{fig:rs_block}}
\end{figure*}

\subsection{Airborne MS-LiDAR Dataset Segmentation}

\textbf{Datasets and Metrics.}
\citep{zhao2021airborne} introduced a large-scale airborne MS-LiDAR dataset, which was captured by a Teledyne Optech Titan MS-LiDAR system. The dataset contains a total of 13 areas, where areas 1-10 are selected as the training areas, and areas 11-13 are selected as the testing areas. Each area covers at least 150,000 $m^{2}$, with an average point density of about 3.6 points/$m^{2}$. The MS-LiDAR data contains three channels with wavelengths of $1,550$ nm (MIR), $1,064$ nm (NIR), and $532$ nm (Green). There are six categories in the dataset: Road, Building, Grass, Tree, Soil, and Powerline, with imbalanced class distributions. The number of road points is more than 30$\times$ the number of powerline points, which makes the dataset more challenging. We took the same data pre-processing (data fusion, normalization, and training/testing sample generation) method described in \citep{zhao2021airborne}. Each area is partitioned into a series of local blocks, each of which contains $4096$ points with six channels (X, Y, Z, MIR, NIR, and Green). Overall Accuracy (OA), mIoU, and average $F_{1}$ score are used for performance evaluation. The model latency is utilized to measure the model efficiency. We also provided the Precision, Recall, and $F_{1}$ score for each category.

\textbf{Performance Comparison.}
The confusion matrix of our MS-LiDAR data segmentation results is shown in Table \ref{tab:semseg_confusion}, where the numbers in the last three rows represent the Precision, Recall, and $F_{1}$ score for each category. As we can see, all the $F_{1}$ scores of all categories except soil are over $85\%$, where the $F_{1}$ scores of Building, Grass, and Tree exceeds $90\%$. Most of the misclassified points are found in areas of soil and roads. This is because the similar geometric characteristics of the soil and roads tend to confuse the network. We also show the comparison results in Table \ref{tab:rs_comparison}. Compared with prior SOTA methods, DTA-Former achieves the best results in terms of OA ($95.4\%$), and average $F_{1}$ score ($88.3\%$), with competitive mIoU ($81.4\%$). It also has the lowest mode latency compared with other efficient Transformer methods such as PointTransformerV3 (PTV3) \citep{PTv3} and DCTNet \citep{lu2024dynamic}.
The visualization results of comparison methods on testing area-11, 12, and 13 are shown in Fig. \ref{fig:rs_comparison_11}, \ref{fig:rs_comparison_12} and \ref{fig:rs_comparison_13}, where the segmentation results of DTA-Former are the most faithful to ground truth. These results demonstrate the excellent performance of DTA-Former in MS-LiDAR point cloud segmentation, exceeding previous SOTA methods. Prediction details of several samples from the testing area-11 are shown in Fig. \ref{fig:rs_block}, with the LTS results in different network stages. It illustrates the excellent prediction results of our method and clearly shows similar geometric characteristics of the soil and roads.
We also explored the robustness of DTA-Former to different input settings in MS-LiDAR data segmentation. Please refer to Section \ref{subsec:ablation} for more details.

\subsection{Aerial DALES Dataset Segmentation}

\textbf{Datasets and Metrics.}
Dayton Annotated LiDAR Earth Scan (DALES) dataset is a comprehensive aerial LiDAR dataset designed for semantic segmentation, introduced by \citep{varney2020dales}. This dataset was collected using a Riegl Q1560 dual-channel system mounted on a Piper PA31 Panther Navajo aircraft. The collection covered 330 $km^{2}$ over Surrey, British Columbia, Canada, with DALES focusing on a 10 $km^{2}$ subset comprising 40 unique tiles, each spanning 0.5 $km^{2}$. The dataset features an average resolution of 50 points per square meter and maintains high accuracy through rigorous calibration and noise filtering. These characteristics make DALES particularly suitable for applications like 3D urban modeling, environmental monitoring, and deep learning algorithm benchmarking.

Each point in the DALES dataset includes 4 attributes: X, Y, Z coordinates, and intensity. 
The dataset encompasses eight semantic classes—Buildings, Cars, Trucks, Poles, Power lines, Fences, Ground, and Vegetation—meticulously hand-labeled with the aid of satellite imagery and a digital elevation model. 
DALES represents one of the largest publicly available aerial LiDAR datasets, providing over 500 million annotated points. 
It is split into training and testing subsets ($70\%$ and $30\%$, respectively) to facilitate robust machine learning evaluations. For a fair comparison, each area is sampled using a 10 $cm$ grid. The 40 areas are split into training and testing sets with roughly a 70/30 percentage. Furthermore, each area is divided into a series of $20 m \times 20 m$ blocks as training/testing samples, where each of them contained 8192 points after sampling. The mIoU and OA are used for performance evaluation.

\begin{table*}[bp]
 \caption{Performance comparison ($\%$) of different methods on the DALES dataset, including OA, mIoU, and latency(ms). \label{tab:dales}
 }
 \centering
 \setlength{\tabcolsep}{25pt}
  \renewcommand{\arraystretch}{1.2}
 \begin{tabular}{l|l|l|l}
  \hline
   {Methods} & OA & mIoU  & Latency(ms)\\
  \hline
  {PointNet++}\citep{qi2017pointnet++}  & 95.7  & 68.3 & 487.1 \\
  {KPConv} \citep{thomas2019kpconv} & 96.9 & 72.4 &125.3\\
  {DGCNN}\citep{wang2019dynamic} & 96.1  & 66.4 &136.2 \\
  {PointCNN}\citep{li2018pointcnn} & 97.2  & 58.4 & - \\
   {SPG}\citep{landrieu2018large}& 95.5 &60.6 &-\\

  {ConvPoint}\citep{boulch2020convpoint}  & 97.2 & 67.4 &-\\
  {PointTransformer}\citep{zhao2021point} & 97.1 &74.9  &468.4  \\

  {SuperCluster}\citep{robert2024scalable}  & - &77.3 &-\\
  {PReFormer}\citep{preformer}  & 92.9 &70.9 &-\\
  {PTV3}\citep{PTv3}  & 96.9 &77.4 &41.9\\
  \hline
  {DTP-Former}   & \textbf{97.2}  & \textbf{78.5} & \textbf{40.7}\\ 

  \hline
 \end{tabular}
\end{table*}

\begin{figure*}[htbp]
  \centering
  \includegraphics[width=0.95\linewidth]{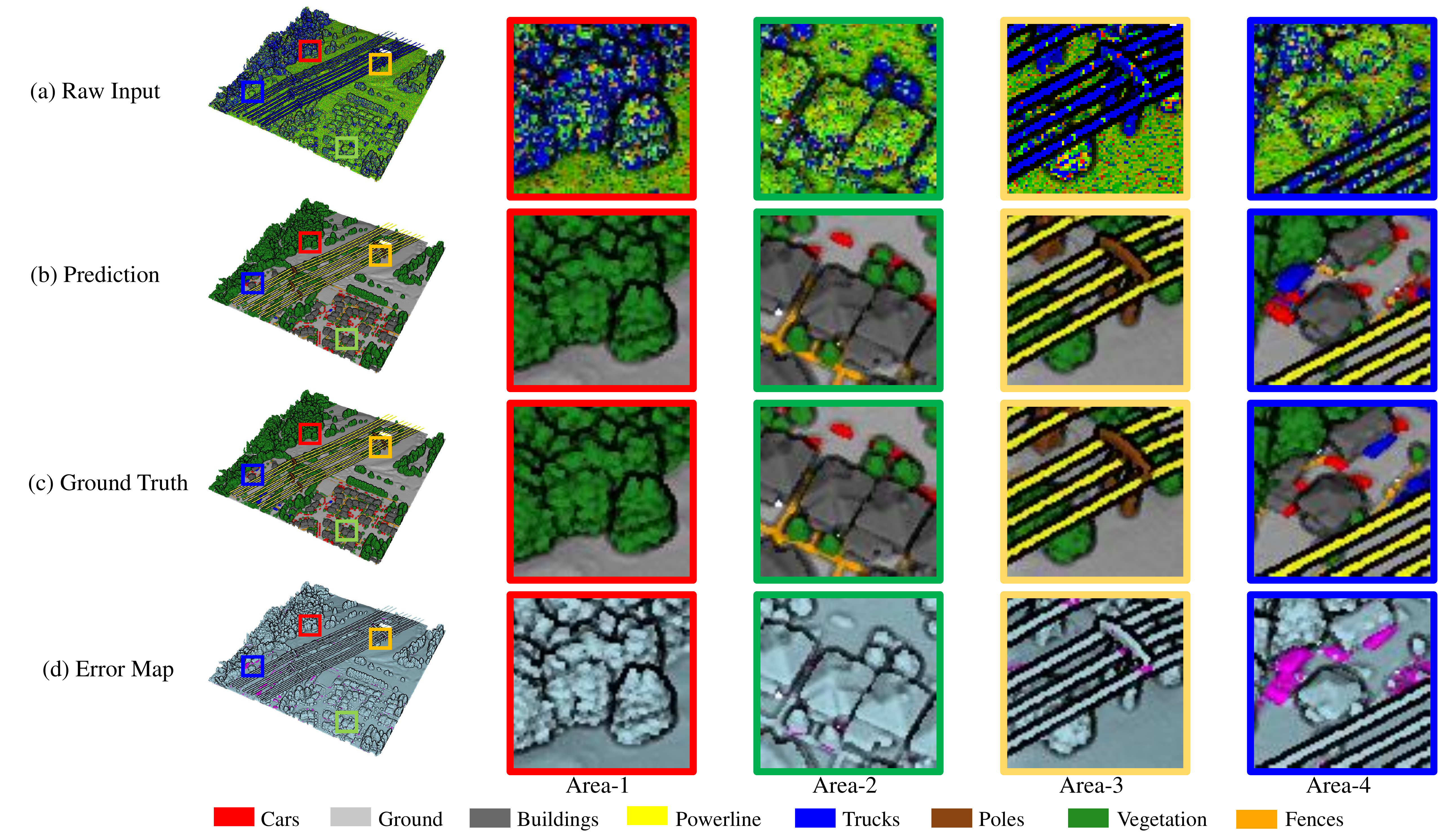}
  \caption{Visualization results of DTA-Former on the DALES dataset, as well as the corresponding error map. The misclassified points in the error map are colored in red.
  \label{fig:dales}}
\end{figure*}

\textbf{Performance Comparison.} The comparison results of DTA-Former with existing related works are shown in Table \ref{tab:dales}. It achieves the highest accuracies (78.5$\%$ of mIoU and 97.2$\%$ of OA) compared with SOTA point cloud segmentation methods such as PTV3 \citep{PTv3} and PReFormer \citep{preformer}. In terms of model efficiency, it also surpasses PReFormer, and achieves the similar latency (40.7ms) with PTV3. In addition, the visualization results of DTA-Former on DALES are shown in Fig. \ref{fig:dales}, as well as the error map. From the results, the prediction results of most areas in the sample shown are faithful to the ground truth, as shown in areas-1, 2, and 3. However, the misclassified points are mainly concentrated in the car and truck categories, as shown in area-4, due to the very similar geometric and reflectivity intensity characteristics of these two categories.

\renewcommand{\arraystretch}{1.2}
\begin{table*}[bp]\color{black}
 \caption{Part segmentation results on the ShapeNet dataset. The highest evaluation score is shown in bold type. \label{tab:partseg_shapenet}
 }
 \setlength{\tabcolsep}{26pt}
  \fontsize{9}{9}\selectfont
 \centering
 \begin{tabular}{llll}
  \hline
   {Methods} & Input Size & mIoU ($\%$) & {Latency (ms)} \\
  \hline
  \multicolumn{4}{l}{Other Learning-based Methods} \\
  \hline
  {PointNet++(MSG)}\citep{qi2017pointnet++} &   2048  & 85.1 & 47.7   \\
  {PointCNN}\citep{li2018pointcnn} &   2048 & 86.1 
 & 252.1   \\
  {DGCNN}\citep{wang2019dynamic}  &   2048 & 85.2  & 96.2   \\
  {DeepGCN}\citep{li2019deepgcns} &   2048 & 86.3  & 105.6    \\
  {RepSurf}\citep{ran2022surface} &   2048 & 86.4  & 117.1    \\
  {Point-PN}\citep{zhang2023starting} &   2048 & \textbf{86.6} & 58.9  \\
  {APES}\citep{wu2023attention} &   2048 & 85.8  & \textbf{24.8}    \\
  \hline
  \multicolumn{4}{l}{Transformer-based Methods} \\
  \hline
  {PointASNL}\citep{yan2020pointasnl} & 2048 & 85.4   & 1023.2  \\
  {PointCloudTransformer}\citep{guo2021pct} &   2048 & 86.4   & 101.1      \\
  {PointTransformer}\citep{zhao2021point}  &  2048 & 86.6   & 560.2    \\
  {GBNet}\citep{qiu2021geometric} &   2048  & 85.9  & 198.4     \\  
   {Stratified Transformer}\citep{lai2022stratified} &   2048  & 86.6   & 1792.6      \\
  {PatchFormer}\citep{zhang2022patchformer} &   2048  & 86.5   & 45.8    \\
  {Li et al.}\citep{li20233d} &   2048  & 86.0   & -    \\
  {Hassan et al.}\citep{hassan2023residual} &   2048  & 86.3 & -   \\
  \hline

  {DTA-Former}    &   2048   &\textbf{86.7}   &\textbf{ 20.7 }     \\

  \hline
 \end{tabular}
\end{table*}

\begin{figure*}[htbp]
  \centering
  \includegraphics[width=0.85\linewidth]{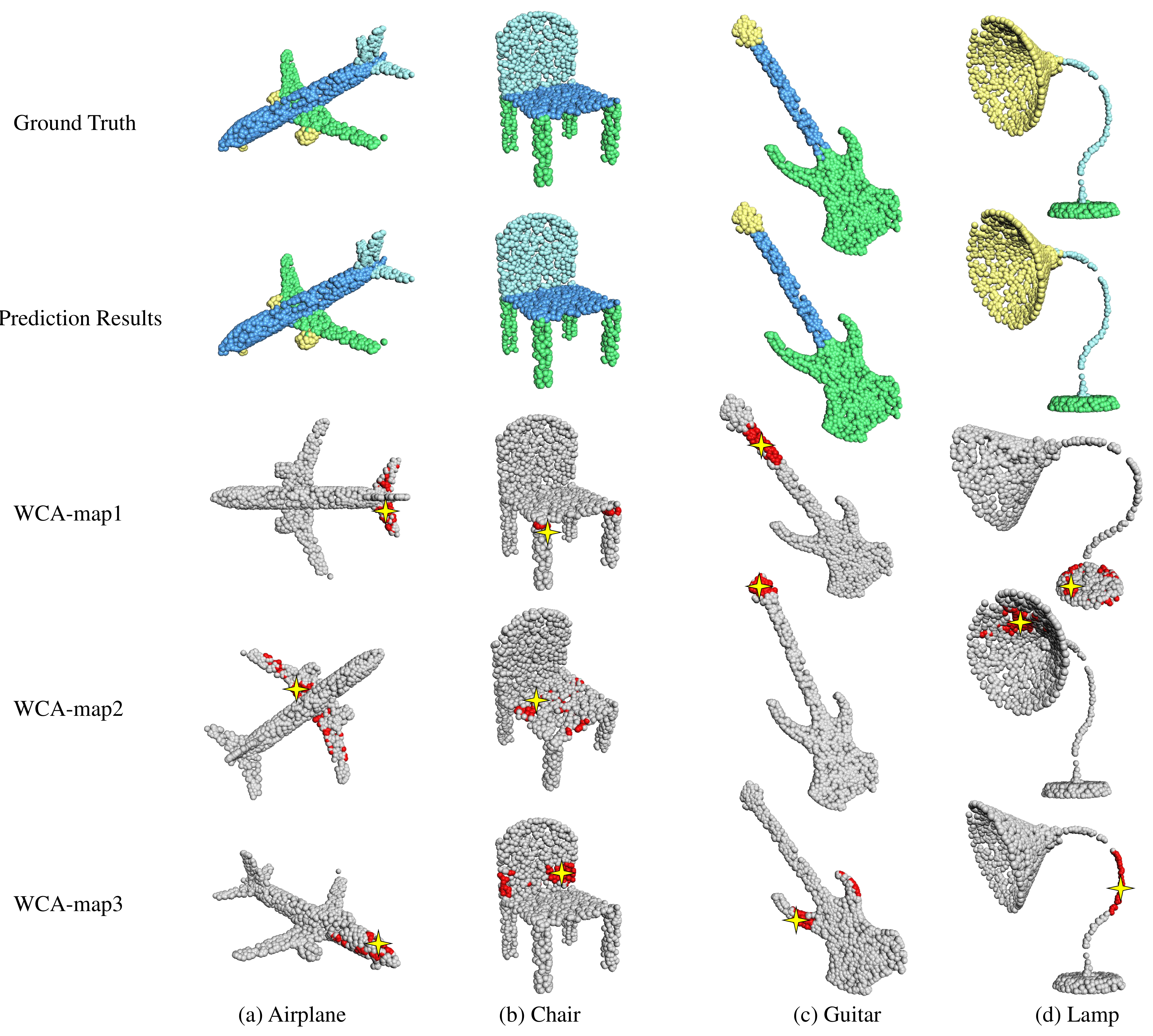}
  \caption{Illustration of part segmentation and WCA-map results on ShapeNet. The prediction results from DTA-Former are faithful to the ground truth. To clearly illustrate the relationship between attention weights and point semantics, we binarized the attention weights from WCA-maps using an appropriate threshold, where high-correlation points are visualized in red, while low-correlation points are displayed in white. The query points are represented by yellow stars.
  \label{fig:wca_result}}
\end{figure*}

\subsection{Point Cloud Part Segmentation}
\textbf{Datasets and Metrics.}
We tested the part segmentation performance of DTA-Former on the ShapeNet dataset to verify the effectiveness of the method and proposed blocks. The dataset contains over $16800$ models with $16$ shape categories and $50$ part labels. Following PointTransformer \citep{zhao2021point}, the original dataset is split into a training set and a testing set, accounting for 80$\%$ and 20$\%$ respectively. Each shape in the dataset has at least two parts. For a fair comparison, each input point cloud was downsampled to $2048$ points with normals by FPS.
The instance-wise mean Intersection over Union (referred to as mIoU in this paper) is used as the evaluation metric for part segmentation. Moreover, the model latency is also used to measure the efficiency of DTA-Former in dense prediction.

\textbf{Performance Comparison.}
Table \ref{tab:partseg_shapenet} shows the comparison results of DTA-Former and existing related methods on the ShapeNet dataset. The benchmarked methods are also divided into two categories: Transformer-based methods and other deep learning-based methods. As measured by mIoU, DTA-Former achieves the best results (86.7$\%$) compared with the prior SOTA methods such as APES \citep{wu2023attention}. With similar accuracy, DTA-Former achieves the lowest latency (20.7ms). These demonstrate the excellent performance of DTA-Former in part segmentation. To illustrate the effectiveness of the proposed blocks, we also show the visualization of part segmentation results in Fig. \ref{fig:wca_result}, as well as the WCA-maps from the DTA block. From the results, our segmentation predictions are faithful to the ground truth. Moreover, the WCA-map visualization reveals a strong correlation between attention weights and point semantics, which illustrates the effectiveness of the proposed DTA in semantic information aggregating.

\subsection{Ablation Study}
\label{subsec:ablation}
Taking DTA-Former as the baseline, we conducted a series of ablation studies on the airborne MS-LiDAR dataset to verify the effectiveness of key blocks in DTA-Former and explore the robustness of the model to different input settings.

\begin{table*}[htbp]\color{black}
 \centering
\setlength{\tabcolsep}{15pt}
 \renewcommand{\arraystretch}{1.2}
 \caption{Ablation results (\%) of key blocks in DTA-Former, which were performed on the airborne MS-LiDAR dataset. $-$ means component removing, and $\rightarrow$ means component changing. 
 }
 \label{tab:ablation}
 \resizebox{\linewidth}{!}{
 \begin{tabular}{lllllll}
  \hline
    \multicolumn{2}{l}{Ablation} &Input Size & Average $F_{1}$ & mIoU & OA  & {Latency (ms)}  \\
 \hline
  \multirow{3}{*}{LTS} & $\rightarrow$Farthest Point Sampling &   4096 & 87.4  & 79.3 & 95.1 & 181.8 \\
  \cline{2-7}
      & $\rightarrow$Random Sampling &   4096 &80.2  & 74.9 & 90.7 &32.6 \\
   \hline
   \multirow{3}{*}{DTA} 
      &$-$ &   4096 & 80.5  & 75.7& 91.9 & \textbf{22.7} \\
   \cline{2-7}
     &$\rightarrow$$k$NN + MLP&   4096& 81.7  & 77.4 & 91.1 & 42.3  \\
  \cline{2-7}
     &WCA$\rightarrow$VCA&   4096& 84.3  & 79.1 & 94.0 & 32.8  \\
    \hline
    \multirow{3}{*}{GFE} 
      &$-$&   4096 & 82.4  & 77.6 & 91.9 & 22.9  \\
  \cline{2-7}
     &$-$ PSA &   4096& 86.6  & 79.0 & 94.2 & 25.4  \\
     \cline{2-7}
     &$-$ CSA &   4096 & 87.0  & 79.7 & 94.5 & 31.5 \\
    \hline
  \multirow{3}{*}{ITR} & $\rightarrow$Trilinear interpolation &   4096 & 83.1  & 78.4 & 92.8 & 36.2 \\
  \cline{2-7}
     &$\rightarrow$Nearest neighbor interpolation &   4096 & 82.5  & 78.2 & 92.7 & 32.1\\
     \cline{2-7}
     &Equipped in U-net &   4096 & 81.5  & 75.3 & 90.8 & 29.3\\
   \hline
   \hline

  \multicolumn{2}{l}{\textbf{DTA-Former} (Baseline)} &   4096 & \textbf{88.3}  & \textbf{81.4} & \textbf{95.4 }& 32.8   \\ 
  \hline
 \end{tabular}}
\end{table*}

\textbf{Learnable Token Sparsification.}
We investigate the effectiveness of the LTS block by comparing it to the commonly used FPS and random sampling approaches. As shown in Table \ref{tab:ablation} Row 2-3, DTA-Former with LTS achieves better results than the other approaches in all three metrics (OA, mIoU, and average $F_{1}$ score), and is over $5\times$ faster than that with FPS in terms of latency. Although LTS is slightly lower than random sampling in terms of inference speed, it outperforms random sampling by 8.1 absolute percentage points in terms of average $F_{1}$ score. This is because random sampling is very sensitive to point cloud density, which leads to its poor performance in real-world LiDAR data processing.

\textbf{Dynamic Token Aggregating.}
A series of ablation experiments were conducted to verify the effectiveness of DTA. We first remove the DTA block from the network, which means the output sparsified tokens are directly fed into the GFE block in each stage. Accordingly, the token reconstruction process is realized by the common nearest interpolation method due to the lack of the WCA-map $WM$. As shown in Table \ref{tab:ablation} Row 4, the performance of the model without DTA drops significantly, where OA drops by 3.5 percentage points, mIoU drops by 5.7 percentage points, and average $F_{1}$ score drops by 7.8 percentage points. Secondly, we replaced DTA with a common local feature aggregation method, $k$NN $+$ MLP. From Table \ref{tab:ablation} Row 5, there are significant drops in both the accuracy and efficiency of the method. These results demonstrate the superiority of DTA. Finally, we replaced the WCA mechanism with a Vanilla Cross-Attention (VCA) one. As shown in Table \ref{tab:ablation} Row 6, we also observe a 1.4, 2.3, and 4.0 drop in OA, mIoU, and average $F_{1}$ score respectively. This suggests that the WCA mechanism has a positive effect on model performance.

\begin{table}[htbp]\color{black}
 \caption{Ablation results (\%) for the numbers of input points. \label{tab:input_ablation}
 }
 \centering
 \setlength{\tabcolsep}{6pt}
 \renewcommand{\arraystretch}{1.2}
 \begin{tabular}{lllll}
  \hline
   {Input Size}  & Average $F_{1}$ & mIoU & OA  & {Latency (ms)}   \\
  \hline
  512 & 79.4  & 68.1 & 88.7 & \textbf{15.2}  \\
  1024  & 81.4  & 71.8 & 92.5 & 17.5   \\
  2048  & 84.7  & 75.4 & 93.0 & 20.7  \\
  4096 & \textbf{88.3}  & \textbf{81.4} & \textbf{95.4}  & 32.8  \\
  8192 & 88.1  & 80.8 & 95.1 & 40.7   \\
  \hline
 \end{tabular}
\end{table}

\begin{figure}[htbp]
  \centering
  \includegraphics[width=\linewidth]{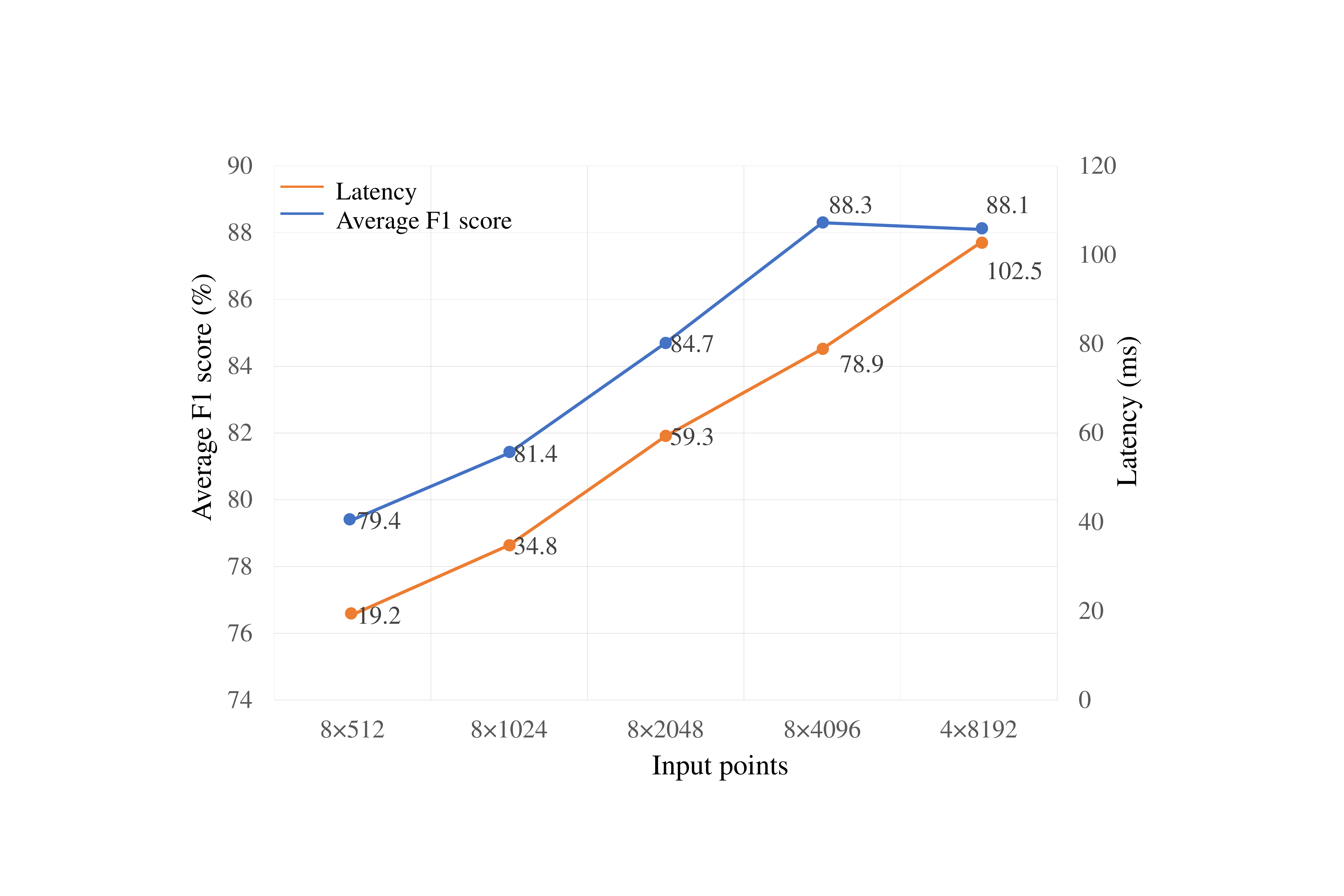}
  \caption{Illustration of model accuracy (represented by average $F_{1}$ score) and efficiency (represented by latency) under different input sizes.
  \label{fig:ab_input_size}}
\end{figure}

\textbf{Global Feature Enhancement.}
Table \ref{tab:ablation} Row 7-9 shows the ablation experiments of the GFE block. The performance of the model without GFE is 91.9$\%$/77.6$\%$/82.4$\%$ in terms of OA/mIoU/average $F_{1}$ score, which is lower than the baseline. To further investigate the contributions of two branches of self-attention mechanisms, we remove them one by one. Specifically, when the point-wise self-attention branch was removed, the three metrics mentioned above dropped by 1.2, 2.4, and 1.7 percentage points respectively. Likewise, when the channel-wise self-attention branch was removed, there is a similar drop in terms of the three metrics. However, the efficiency of the model without point-wise self-attention is higher than that without channel-wise one, which is measured by lower latency. These results demonstrate that the two branches of self-attention mechanisms contribute equally to the model performance in our network, while the channel-wise self-attention mechanism is faster than the point-wise one.

\textbf{Iterative Token Reconstruction.}
Table \ref{tab:ablation} Row 10-12 shows the ablation studies of ITR. We first replaced the ITR block with the common trilinear and nearest neighbor interpolations, respectively. From the results, we can see that the proposed ITR block outperforms the aforementioned two kinds of interpolation methods, demonstrating the benefits of the accurate token-wise semantic relationship captured by DTA. Instead of the W-net architecture, we also explored the segmentation performance of the proposed ITR in the U-net architecture, while keeping other modules unchanged. As shown in Table \ref{tab:ablation} Row 12, the performance of the U-net with ITR drops significantly compared with our W-net architecture. In the U-net architecture, the feature maps in the encoder need to undergo further feature extraction and reconstruction to obtain the corresponding feature maps in the decoder. Therefore, the feature relationships in the decoder cannot be accurately described by the corresponding WCA-maps established by the DTA block in the encoder. In contrast, our W-net segmentation network allows the WCA-maps in the DTA block to be promptly fed back to the sparsified tokens in the same stage for reconstruction. This ensures the timeliness of feature interaction and reconstruction.

\begin{table}[htbp]\color{black}
 \caption{Results (\%) comparison of input data with different channel combinations on the airborne MS-LiDAR dataset.} \label{tab:input_channel}
 \centering
 \setlength{\tabcolsep}{6pt}
 \renewcommand{\arraystretch}{1.2}
 \begin{tabular}{lllllll}
  \hline
   \multicolumn{4}{l}{Input} & \multirow{2}{*}{Average $F_{1}$} &  \multirow{2}{*}{mIoU} &  \multirow{2}{*}{OA}   \\
   \cline{1-4}
   XYZ &MIR & NIR &Green & & & \\
   \hline

   \checkmark & - & \checkmark & \checkmark &84.2 &75.1 &92.7  \\

   \checkmark &\checkmark & - & \checkmark  &78.1 &69.0 &90.5\\

   \checkmark & \checkmark & \checkmark & - &83.1 &74.9 &92.1 \\

   \checkmark & - & - & - &71.8 &62.2 &86.9 \\
   
   \checkmark & \checkmark & \checkmark & \checkmark & \textbf{88.3}  & \textbf{81.4} & \textbf{95.4}\\
  \hline
 \end{tabular}
\end{table}

\textbf{Sensitivities of input settings.} We investigated the robustness of DTA-Former to different input settings of the airborne MS-LiDAR data. Table \ref{tab:input_ablation} and Fig. \ref{fig:ab_input_size} show the accuracy and efficiency of the model under the situations of different numbers $N$ of input points, where $N$ is varied from 8192, to 4096, 2048, 1024, and 512. From the results, the average $F_{1}$ score, mIoU, and OA increase with $N$ until it reaches 4096, while the latency of the model also gradually increases. We also explored the contributions of different channels in the airborne MS-LiDAR data. As shown in Table \ref{tab:input_channel}, we conducted ablation studies on input channels by removing one of the three channels (MIR, NIR, and Green) at a time, assessing the corresponding changes in performance. Compared with the baseline, the removals of MIR and Green channels caused slight yet similar drops in accuracy. In contrast, removing the NIR channel leads to a significant decline in model performance, with the average $F_{1}$ score and mIoU dropping by more than 10\%. Finally, we removed all these three channels and only kept the three geometric coordinates of XYZ as input, which led to a further drop in model performance. The results demonstrate that each of the three channels plays a role in enhancing the segmentation performance of our model, with the NIR channel making the most significant contribution.

\section{Conclusion}
\label{sec:conclusion}
In this work, we proposed an efficient point Transformer network, named DTA-Former for LiDAR point cloud segmentation, consisting of several novel blocks such as LTS, DTA, and ITR. Instead of the common U-net design, DTA-Former has a novel W-net architecture for effective Transformer-based feature learning. It consists of two stages, where each stage is composed in a sequence of LTS, DTA, GFE, and ITR blocks. Specifically, the LTS block calculates a decision score for each input token by considering its both local and global information, followed by token sparsification with a differentiable top-H selection method. From the ablation study, the proposed LTS outperforms the commonly used sampling approaches such as FPS in terms of both accuracy and efficiency. The sparsified tokens are then fed into the DTA block for feature aggregation at a global scale. In the DTA block, a novel WCA is proposed to fully explore the semantic relationship between the sparsified token set and the original token set. Therefore, instead of a local region, DTA allows each sparsified token to perceive semantic information across the global context, expanding the effective receptive field while mitigating information loss which is common in traditional pooling operations. Following the DTA block, a dual-attention Transformer-based GFE block is employed to enhance the ability of the model in long-range dependency modeling. Finally, the ITR block is placed at the end of each stage, mapping the sparsified tokens back into the original token set by the WCA-map generated in the DTA block. From the ablation experiments, compared with the U-net \citep{qi2017pointnet++}, the proposed W-net is more suitable for Transformer-based feature learning. Extensive experiments on the airborne MS-LiDAR semantic segmentation dataset \citep{zhao2021airborne} and DALES dataset \citep{varney2020dales} demonstrate that DTA-Former outperforms previous SOTA methods in LiDAR point cloud segmentation, in terms of both algorithm accuracy and efficiency.

\printcredits

\section*{Declaration of competing interest}
The authors declare that they have no known competing financial interests or personal relationships that could have appeared to influence the work reported in this paper.

\section*{Acknowledgements}
This work was supported in part by the Natural Sciences and Engineering Research Council of Canada (NSERC) under the Discovery Grant No. RGPIN-2022-03741 and Grant No. RGPIN-2019-06744. The first author was sponsored in part by the Chinese Scholarship Council under Grant No. 202106830030.

\bibliographystyle{cas-model2-names}
\bibliography{egbib}

\end{document}